\documentclass{article}
\usepackage{spconf,amsmath,amssymb,graphicx}

\def\RR{\mathbb R}
\title{A Study of Shape Modeling Against Noise}
\name{Cheng Long, Adrian Barbu}
\address{Florida State University\\
    Department of Statistics\\
    Tallahassee, FL 32306}

\begin{document}
%
\maketitle
\begin{abstract}
Shape modeling is a challenging task with many potential applications in computer vision and medical imaging. 
There are many shape modeling methods in the literature, each with its advantages and applications. However, many shape modeling methods have difficulties handling shapes that have missing pieces or outliers. 
In this regard, this paper introduces shape denoising, a fundamental problem in shape modeling that lies at the core of many computer vision and medical imaging applications and has not received enough attention in the literature.
The paper introduces six types of noise that can be used to perturb shapes as well as an objective measure for the noise level and for comparing methods on their shape denoising capabilities.
Finally, the paper evaluates seven methods capable of accomplishing this task, of which six are based on deep learning, including some generative models.
\end{abstract}
\begin{keywords}
shape modeling, shape denoising, object segmentation\vspace{-2mm}
\end{keywords}
%
\vspace{-2mm}
\section{Introduction}
\label{sec:intro}
\vspace{-3mm}

The analysis of shape has attracted attention for decades since shape is the most basic visual feature of an object. 
Due to external circumstances, sometimes one cannot obtain any other information such as the color or texture of an object besides the shape. 
Moreover, other features of an object can be more easily affected by environmental factors. 
For example, an object's appearance depends on the illumination direction and light color. 
Hence, shape can be considered a more stable feature of an object, and this is one reason why shape analysis plays an essential role in many computer vision tasks. 
This paper investigates the problem of restoring object shapes when the shape has been distorted by an external factor, a task that can also be called 'shape denoising'.

To denoise shapes, one needs to use an appropriate shape representation that is flexible enough for this purpose. 
In earlier studies, different methods were designed to extract shape features such as moments \cite{flusser1993pattern}, shape context \cite{belongie2002shape}, curvature \cite{jalba2006shape}, and mathematical morphology \cite{pitas1992morphological}.

There are different ways to represent shape spaces. 
Kendall \cite{kendall1984shape} considers shapes as k-tuples of points in $\RR^d$. Under this notion of shape space, \cite{cootes1995active} proposed the Active Shape Model, which performs a PCA on the points of the aligned shapes of the training objects. All shapes are represented using the same number of points, and these points correspond to each other between objects.

In this work, the shape will be represented by a binary image, which is a form of a level set representation.
Considering a shape as a binary image allows many generative models to be applied to shape modeling, such as Restricted Boltzmann Machines (RBM) \cite{smolensky1986information}, Deep Boltzmann Machines (DBM) \cite{salakhutdinov2009deep}, Centered Convolutional Deep Boltzmann Machines (CCDBM) \cite{yang2021centered} and Energy Based Models (EBM) \cite{pang2020learning}. 
Moreover, deep learning based methods for semantic segmentation can also be used for shape modeling, e.g. U-Net \cite{ronneberger2015u}, DeepLabv3+ \cite{chen2018encoder}, masked autoencoder (MAE) \cite{he2021masked}, etc.

\vspace{-5mm}
\subsection{Related Work}
\label{sec:related}
\vspace{-2mm}

The focus of this study is to evaluate and compare the performance of different modeling methods for handling shapes represented by binary images, corrupted by different types of noise. 
There is a paucity of literature focusing on this topic. However, there is some recent literature containing works about shape modeling.

 A recent study about 2D shape modeling \cite{yang2021centered} introduces convolutions into a DBM-based shape model. 
The goal of their work is to generate realistic shapes that are different from all training shapes. 
In their study, the resulting images are generated from test images without any noise. The quality of the generated images is judged subjectively. This evaluation process can not directly describe the shape modeling capability of the proposed method. 
If the resulting shape is similar to the test shape that has been used for initialization, it only proves that the model can generate results similar to the initial shapes, rather than the ability to represent the entire shape space correctly.
If the resulting shape is quite different from the initialization, it is difficult to say whether it is a good shape belonging to the shape space being modeled. 
In our work, the aim is to recover the object shape from different kinds of noisy perturbations, which allows us to use an objective evaluation criterion: the recovered shape must be close to the original (unseen) shape. 
In \cite{yang2021centered}, all models used for comparison are based on Restricted Boltzmann Machines. Our work expands the scope of comparison by adding more methods that include both classical ones such as the Active Shape Model(ASM) and recent ones such as the Energy-based Model(EBM) \cite{pang2020learning}.

In the field of recovering shapes from noisy images, the Active Shape Model (ASM) \cite{cootes1995active} learns a Point Distribution Model (PDM) from training sets of correctly labeled images with point correspondences, then exploits the linear formulation of the PDM in an iterative search procedure to recover the original shape from a noisy test image. 
In their study, the authors are more focused on how the parameter of each shape feature influences the final shape. In our work, we pay more attention to the quality of recovered shape and use the IoU (Intersection over Union) between the resulting shape and corresponding ground truth shape for evaluation. 
More importantly, as we will see in our experiments, the method only works in some cases and with a good initialization. Our works construct many types of noise for a more challenging and realistic evaluation. 

In addition to external noise or occlusion, the variation in viewing angles will also increase the shape modeling difficulty. \cite{bryner2014bayesian} proposed an elastic, affine-invariant shape model to segment images of objects subject to perspective skew. Our work focuses on external shape noise, that needs to be eliminated from objects viewed from a certain range of angles, which should be the same for training and testing.

Since recovering shape from noisy shapes can be seen as a degenerate version of semantic segmentation, the literature in semantic segmentation is related to our work. 
 The Fully Convolutional Network (FCN) \cite{long2015fully} converts the classification network \cite{krizhevsky2012imagenet} into a fully convolutional network by replacing all the fully connected layers with convolutions. 
FCN has two advantages: it can process input images of any size without losing spatial information and thus reduce computation costs. 
The U-Net \cite{ronneberger2015u} adds a decoder that is symmetrical to encoder based on the FCN architecture and concatenates feature maps from the same level of the encoder and decoder before performing convolutions. 
DeepLab \cite{chen2014semantic} adopts atrous convolution and fully connected conditional random fields (CRF). 

In the process of improvement, including DeepLabv2 \cite{chen2017deeplab}, DeepLabv3 \cite{chen2017rethinking} and DeepLabv3+ \cite{chen2018encoder}, CRF is deprecated, an atrous spatial pyramid pooling (ASPP) module was proposed.

\vspace{-5mm}
\section{The Shape Denoising Problem}
\label{sec:problem}
\vspace{-2mm}

\begin{figure}[t]
\begin{minipage}[b]{0.32\linewidth}
\vspace{-3mm}
  \centering
  \centerline{\includegraphics[width=2.5cm]{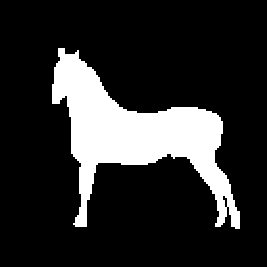}}
  \centerline{(a) Original shape}\medskip
\end{minipage}
\hfill
\begin{minipage}[b]{0.32\linewidth}
  \centering
  \centerline{\includegraphics[width=2.5cm]{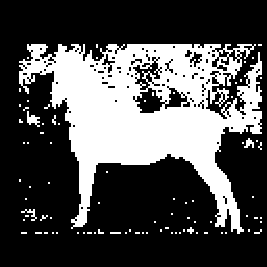}}
  \centerline{(b) Noisy shape}\medskip
\end{minipage}
\hfill
\begin{minipage}[b]{0.32\linewidth}
  \centering
  \centerline{\includegraphics[width=2.5cm]{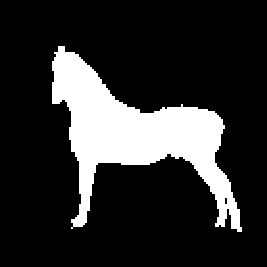}}
  \centerline{(c) Denoised shape}\medskip
\end{minipage}
\vskip -6mm
\caption{Shape denoising example. The noisy shape (b) has been obtained from the original shape (a) by a noise inducing process such as those described in Section \ref{sec:noise}. A shape denoising method is used to obtain the denoised shape (c).}
\label{fig:denoising process}
\vspace{-6mm}
\end{figure}

Shape denoising is the process of removing the noise from a shape, with the goal of obtaining a shape as close to the original shape as possible. Figure \ref{fig:denoising process} illustrates an example of shape denoising.

In this chapter we introduce the shape representation that will be used in our study and six different types of noise that can be used to perturb shapes.

\vspace{-4mm}
\subsection{Shape Representation}
\label{sec:representation}
\vspace{-1mm}
In this work, shapes will be represented as binary images of a certain size($128 \times 128$ in our experiments). 
All the shape images have black background and white foreground. All foregrounds are centered in the image and have approximately similar sizes, obtained in an alignment step described below.

\noindent{\bf Shape alignment.} For each binary image $I$, let $I(x,y) \in \{0,1\}$ be the intensity at location $(x,y)$. Let $C_1$ be the foreground region, $C_1 = \{(x,y)|I(x,y) = 1\}$ and $C_0$ the background, $C_0 = \{(x,y)|I(x,y) = 0\}$. The center of mass of the foreground is
\vspace{-5mm}
\begin{equation}
		(\bar{x}, \bar{y}) = \left(\frac{1}{N} \sum_{i=1}^{N} x_i, \frac{1}{N} \sum_{i=1}^{N} y_i\right),
\vspace{-3mm}
\end{equation}
where $N = |C_1|$, $(x_i, y_i) \in C_1, i=\overline{1,N}$.
Let $d(x_i,y_i)$ be the distance between point $(x_i,y_i)\in C_1$ and the center point $(\bar{x}, \bar{y})$,
\vspace{-2mm}
\begin{equation}
		d(x_i,y_i) = \left\lVert (x_i-\bar{x}, y_i - \bar{y}) \right\rVert_2 .
\vspace{-1mm}
\end{equation}
Let $p_{80}$ be the $80\%$ percentile of the set $D = \{ d(x_i,y_i)|i=\overline{1,N} \}$. 
The binary image is then rescaled using the scale factor   $40/p_{80}$ using bicubic interpolation. The obtained grayscale image is then thresholded to obtain a binary image again, denoted by $I'$. 
Let $C_1' = \{(x,y)|I'(x,y) = 1\}$ and $C_0' = \{(x,y)|I'(x,y) = 0\}$ be its foreground and background region respectively.
The new center point of the foreground is
\vspace{-6mm}
\begin{equation}
		(\bar{x}', \bar{y}') = \left(\frac{1}{N'} \sum_{i=1}^{N'} x_i, \frac{1}{N'} \sum_{i=1}^{N'} y_i\right)
\vspace{-4mm}
\end{equation}
where $N' = |C_1'|$, $(x_i, y_i) \in C_1', i=\overline{1,N'}$. The image $I'$ is then cropped to obtain an image of size $128 \times 128$ centered at $(\bar{x}', \bar{y}')$. Padding is used if necessary.

\vspace{-4mm}
\subsection{Introducing Noise in Shapes}
\label{sec:noise}
\begin{figure}[t]
\vspace{-3mm}
\centering
\begin{tabular}{cccc}
\hspace{-2mm}\includegraphics[height=1.9cm]{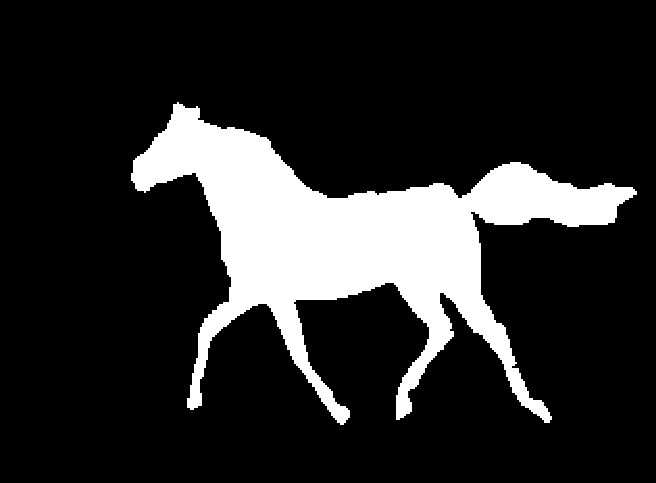}
&\includegraphics[height=1.9cm]{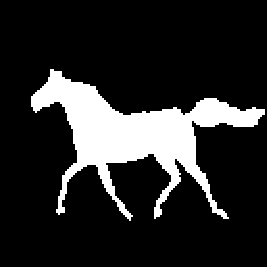}
  &\includegraphics[height=1.9cm]{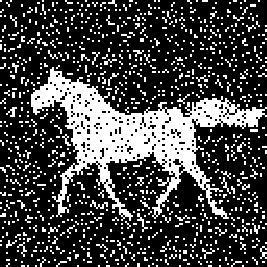}
  &\includegraphics[height=1.9cm]{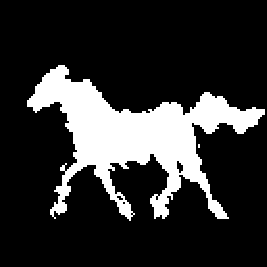}\\
\hspace{-2mm}(a) Original &(b) Aligned&(c) Salt &(d) Circle\\
\hspace{-2mm}\includegraphics[height=1.9cm]{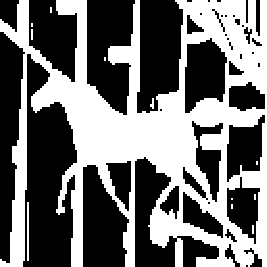}
&\includegraphics[height=1.9cm]{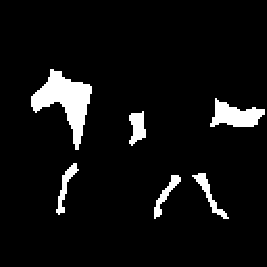}
&\includegraphics[height=1.9cm]{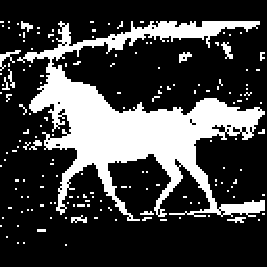}
&\includegraphics[height=1.9cm]{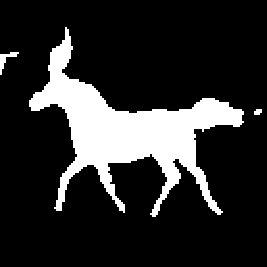}\\
\hspace{-2mm}(e) Real&(f) Occlusion&(g) Thr. prob.&(h) Detection\\
\end{tabular}
\vskip -3mm
\caption{Illustration of shape alignment and the six types of shape noise introduced in this work. (a) Shape before alignment, (b) Shape after alignment, (c) Salt and pepper noise, (d) Circle noise, (e) Real image noise, (f) Occlusion noise, (g) Thresholded probability noise, (h) Detection image noise.}
\label{fig:six noise}
\vspace{-5mm}
\end{figure}

\vspace{-1mm}
We will introduce six types of noise that
could be used for perturbing shapes to evaluate the capability of different shape denoising methods in recovering the original shape.
The six types of noise are: salt and pepper noise, circle noise, real image noise, occlusion noise, detection image noise and thresholded probability noise, illustrated in Fig. \ref{fig:six noise}.

\noindent \textbf{Salt and pepper noise} is obtained by flipping each pixel to its opposite value with a probability $p$. For this reason it could also be called Bernoulli noise. The flipping probability $p$ is used to control the noise level.

\noindent \textbf{Circle noise} is obtained by adding semicircles or punching holes at random locations on the boundary between the foreground and the background. The radius $r$ of the semicircles or holes is used to control the noise level. For each noise image, the radius is fixed.

\noindent \textbf{Real image noise.} Real images are binarized by thresholding with various thresholds and the obtained binary image is used to replace the background pixels.

\noindent \textbf{Occlusion noise.} Part of the shape is occluded.

\noindent \textbf{Thresholded probability noise.} For any binary segmentation $M$, the corresponding original color image $I$ is used to generate a probability map as follows.
Let $I(x,y) \in \{0,1,...,255\}^3$ be the image intensity at location $(x,y)$. 
Denote $N$ as the number of pixels in the color image, the intensities of all pixels can be represented by a $N$ points in $\RR^3$. 
The $N$ points are clustered into $k$ clusters using k-means clustering, obtaining cluster indices for all pixels $L \in \{1,2,...,k\}^N$. Let $C_1=\{(x,y), M(x,y)=1\}$ be the foreground region and $C_0=\{(x,y), M(x,y)=0\}$ the background region. For each cluster $i \in \{1,2,..k\}$, the number of pixels to the foreground region or the background are computed:
\vspace{-3mm}
\begin{equation}
		N_{i1} = | \{(x,y)|L(x,y)=i \land (x,y) \in C_1\} |
\vspace{-3mm}
\end{equation}
\begin{equation}
		N_{i0} = | \{(x,y)|L(x,y)=i \land (x,y) \in C_0\} |
\vspace{-2mm}
\end{equation}
Then the probability map of color image $I$ can be computed as follows:
\vspace{-4mm}
\begin{equation}
		P(x,y) = \frac{N_{j1}}{N_{j1} + N_{j0}}, \text{ where $j = L(x,y)$.}
\vspace{-4mm}
\end{equation}
Then binary noisy shapes are obtained by applying different thresholds to the probability map.

\noindent \textbf{Detection noise.} This is the noise introduced during the segmentation of an object from a color or grayscale image. 
In our experiments, we used a trained Fully Convolutional Network classifier to predict the foreground/background label for the pixels of an image.

\vspace{-5mm}
\subsection{Shape Denoising Evaluation}
\vspace{-2mm}

The capability of a shape modeling method to extract the object shape from a noisy input will be measured using the \textbf{Intersection over Union (IoU)}, also known as the Jaccard Index. 
The IoU is a measure of similarity between two sets, with a range from 0 to 1, with larger values indicating higher similarity. 
For a binary image $I$, let $C_I$ be the foreground  region, $C_I = \{(x,y)|I(x,y) = 1\}$. 
For two binary images $A,B$ the IoU is defined as:
\vspace{-3mm}
\begin{equation}
IoU(A,B)\hspace{-0.5mm} = \hspace{-0.5mm}\frac{|C_A \cap C_B|}{|C_A \cup C_B|} 
\hspace{-0.5mm}=\hspace{-0.5mm} \frac{|C_A \cap C_B|}{|C_A|\hspace{-0.5mm} +\hspace{-0.5mm} |C_B|\hspace{-0.5mm} - \hspace{-0.5mm}|C_A \cap C_B|}
\label{eq:iou}
\vspace{-3mm}
\end{equation}
This measure will be used to evaluate the IoU between the denoised shape obtained by a denoising method and the ground truth shape on which the noise was applied to.

\vspace{-5mm}
\section{Evaluation of Shape Denoising Methods}
\label{sec:evaluation}
\vspace{-3mm}

We evaluate seven shape denoising methods (ASM \cite{cootes1995active}, DBM \cite{salakhutdinov2009deep}, CDBM \cite{yang2021centered}, EBM \cite{pang2020learning}, U-Net \cite{ronneberger2015u}, Deeplabv3+ \cite{chen2018encoder}, MAE \cite{he2021masked}) for denoising shapes corrupted by the six types of noise introduced in Section \ref{sec:noise}. 
The criterion we use to estimate the quality of the denoising result is the IoU \eqref{eq:iou}.

\noindent \textbf{The Weizmann Horse dataset} \cite{borenstein2004combining} contains 327 horse images and their corresponding mask images. All mask images were aligned as described in section \ref{sec:representation} to have all shapes centered and of approximately the same size. 
The aligned images were resized to $128 \times 128$.
From the 327 aligned images,159 images were randomly selected as the training set $S_{train}^{clean}$ and the other 168 images as the test set $S_{test}^{clean}$.

\noindent \textbf{The Caltech-UCSD Birds 200 dataset} \cite{WelinderEtal2010} contains photos of 200 bird species. We use 417 images of seven Flycatcher species in our experiment. The images had ground truth segmentations and were aligned as described in section \ref{sec:representation} and resized to $128 \times 128$. From the 417 aligned images, 207 images were randomly selected as the training set $S_{train}^{clean}$ and the other 210 images as the test set $S_{test}^{clean}$. 

\noindent \textbf{Training Sets.} For each image in the training set $S_{train}^{clean}$, noisy versions were generated as follows. \textbf{Salt and pepper noise} was generated using 16 flipping probabilities $p\in \{0,0.01,0.02,...,0.15\}$. \textbf{Circle noise} was generated using 11 radii  $r\in \{0,1,2,...,10\}$. \textbf{Real image noise} was generated by randomly selecting a $128\times 128$ size random patches from a randomly image of the PASCAL VOC 2012 Dataset. The thresholds to generate binary noise images are $\{10/255,20/255,30/255,...,250/255\}$.

\noindent \textbf{Test Sets.} For each image in the test set $S_{test}^{clean}$, we use the above methods to generate four types of noisy images. 
A trained FCN was used to generate detection image noise. 
We organized all noisy shapes by their IoU with the original shape. 
For each IOU level (e.g. 0.5-0.6, 0.6-0.7, ...) we randomly selected at most 1000 shapes for testing.

 \begin{table}[t]
	\begin{center}
		\caption{Performance (IoU) of the methods evaluated on test sets for the Weizmann Horse dataset.}
		\label{tab:compare horse}
\scalebox{0.9}{
		\begin{tabular}{|l|c|c|c|c|c|c|c|}
			\hline
			{Input IoU}&ASM& DBM& CDBM& EBM& U-Net& Deeplabv3+& MAE\\
			\hline
			\multicolumn{8}{l}{Salt and Pepper Noise}\\
			\hline
			{0.5-0.6}& 0.476& 0.677& 0.833& 0.881& {\bf 0.966}& 0.926& 0.963\\
			\hline
			{0.6-0.7}& 0.564& 0.704& 0.873& 0.883& {\bf 0.976}& 0.934& 0.973\\
			\hline
			{0.7-0.8}& 0.616& 0.717& 0.893& 0.887& {\bf 0.983}& 0.940& 0.982\\			
			\hline
			{0.8-0.9}& 0.629& 0.724& 0.896& 0.893& 0.988& 0.944& {\bf 0.989}\\			
			\hline
			{0.9-1}& 0.653& 0.720& 0.889& 0.895& 0.992& 0.941& {\bf 0.996}\\	
			\hline
			\multicolumn{6}{l}{Circle Noise}\\
			\hline
			{0.5-0.6}& 0.558& 0.598& 0.588& 0.637& 0.818& 0.751& {\bf 0.848}\\
			\hline
			{0.6-0.7}& 0.600& 0.644& 0.657& 0.705& 0.865& 0.799& {\bf 0.888}\\
			\hline
			{0.7-0.8}& 0.645& 0.685& 0.723& 0.767& 0.897& 0.833& {\bf 0.914}\\			
			\hline
			{0.8-0.9}& 0.667& 0.714& 0.797& 0.824& 0.913& 0.865& {\bf 0.930}\\			
			\hline
			{0.9-1}& 0.667& 0.726& 0.882& 0.884& 0.952& 0.915& {\bf 0.963}\\
			\hline
			\multicolumn{6}{l}{Real Image Noise}\\
			\hline
			{0.5-0.6}& 0.492& 0.648& 0.660& 0.805& 0.933& 0.868& {\bf 0.942}\\
			\hline
			{0.6-0.7}& 0.565& 0.683& 0.730& 0.843& 0.954& 0.890& {\bf 0.963}\\
			\hline
			{0.7-0.8}& 0.625& 0.710& 0.790& 0.869& 0.970& 0.910& {\bf 0.978}\\			
			\hline
			{0.8-0.9}& 0.644& 0.718& 0.839& 0.886& 0.980& 0.922& {\bf 0.987}\\
			\hline
			{0.9-1}& 0.657& 0.721& 0.881& 0.895& 0.990& 0.934& {\bf 0.995}\\
			\hline
			\multicolumn{6}{l}{Occlusion Noise}\\
			\hline
			{0.5-0.6}& 0.285& 0.636& 0.630& 0.706& {\bf 0.849}& 0.793& {\bf 0.852}\\
			\hline
			{0.6-0.7}& 0.361& 0.655& 0.670& 0.734& {\bf 0.875}& 0.815& {\bf 0.874}\\
			\hline
			{0.7-0.8}& 0.467& 0.665& 0.715& 0.761& {\bf 0.885}& 0.829& {\bf 0.886}\\			
			\hline
			{0.8-0.9}& 0.559& 0.687& 0.775& 0.801& 0.909& 0.853& {\bf 0.914}\\			
			\hline
			{0.9-1}& 0.643& 0.715& 0.858& 0.870& {\bf 0.961}& 0.905& {\bf 0.964}\\	
			\hline
			\multicolumn{6}{l}{Thresholded Probability Noise}\\
			\hline
			{0.5-0.6}& 0.438& 0.623& 0.640& 0.719& 0.839& 0.788& {\bf 0.845}\\
			\hline
			{0.6-0.7}& 0.507& 0.664& 0.712& 0.776& 0.865& 0.813& {\bf 0.868}\\
			\hline
			{0.7-0.8}& 0.577& 0.701& 0.773& 0.823& 0.888& 0.846& {\bf 0.893}\\			
			\hline
			{0.8-0.9}& 0.622& 0.714& 0.822& 0.857& 0.915& 0.884& {\bf 0.917}\\			
			\hline
			{0.9-1}& 0.674& 0.732& 0.863& 0.884& 0.939& 0.905& {\bf 0.941}\\
			\hline
			\multicolumn{6}{l}{Detection Image Noise}\\
			\hline
			{0.5-0.6}& 0.288& 0.632& 0.617& 0.728& {\bf 0.833}& 0.749& {\bf 0.812}\\
			\hline
			{0.6-0.7}& 0.486& 0.673& 0.656& 0.743& {\bf 0.797}& 0.743& {\bf 0.803}\\
			\hline
			{0.7-0.8}& 0.621& 0.672& 0.728& 0.778& {\bf 0.805}& 0.786& {\bf 0.822}\\			
			\hline
			{0.8-0.9}& 0.639& 0.701& 0.810& 0.843& {\bf 0.878}& 0.856& {\bf 0.878}\\			
			\hline
			{0.9-1}& 0.638& 0.742& 0.869& 0.886& {\bf 0.930}& 0.911& {\bf 0.931}\\	
			\hline
		\end{tabular}
	}
	\end{center}
	\vspace {-10mm}
\end{table}
\noindent \textbf{Training details.} We used the clean shapes in $S_{train}^{clean}$ to train the models for ASM, DBM, and Convolutional Deep Boltzmann Machines(CDBM). The EBM, U-Net, Deeplabv3+ and MAE were trained with clean shapes as well as shapes perturbed by all types of noise except the detection noise. 

\noindent \textbf{Results.} The comparison of all methods on the test sets are displayed in Tables \ref{tab:compare horse} and \ref{tab:compare Flycather}. All results that are not significantly worse than the best result in a one-sided paired t-test with $\alpha = 0.05$ are bolded. ASM was not evaluated for the Flycatcher data because there were no manually labeled keypoints available to obtain point correspondences.

\noindent \textbf{Discussion.} The experiments reveal that MAE and U-Net are the best shape denoising methods we evaluated for all six types of noise. 
DeepLabv3+ is the third best shape denoising method for the six noise types in most situations. 
EBM outperforms CDBM on all six noise types, especially when dealing with real image noise.
In terms of the difficulty of the various types of noise, the salt and pepper noise is the easiest to deal with, followed by real image noise.
Circle noise and occlusion noise are more challenging than the above two, especially when the noise level is high. 
The most challenging noises among these six are the thresholded probability noise and detection image noise.

 \begin{table}[t]
	\begin{center}
		\caption{Performance (IoU) of the methods evaluated on test sets(Flycatcher in Caltech-UCSD Birds 200 dataset).}
		\label{tab:compare Flycather}
\scalebox{0.9}{
		\begin{tabular}{|l|c|c|c|c|c|c|}
			\hline
			{Input IoU}& DBM& CDBM& EBM& U-Net& Deeplabv3+& MAE\\
			\hline
			\multicolumn{7}{l}{Salt and Pepper Noise}\\
			\hline
			{0.5-0.6}& 0.627& 0.814& 0.891& {\bf 0.977}& 0.942& 0.975\\
			\hline
			{0.6-0.7}& 0.635& 0.811& 0.893& {\bf 0.983}& 0.946& 0.982\\
			\hline
			{0.7-0.8}& 0.643& 0.816& 0.901& {\bf 0.988}& 0.952& {\bf 0.988}\\			
			\hline
			{0.8-0.9}& 0.645& 0.812& 0.898& 0.987& 0.953& {\bf 0.992}\\			
			\hline
			{0.9-1}& 0.656& 0.819& 0.909& 0.988& 0.955& {\bf 0.996}\\	
			\hline
			\multicolumn{6}{l}{Circle Noise}\\
			\hline
			{0.5-0.6}& 0.434& 0.570& 0.542& 0.685& 0.684& {\bf 0.821}\\
			\hline
			{0.6-0.7}& 0.556& 0.662& 0.662& 0.799& 0.778& {\bf 0.869}\\
			\hline
			{0.7-0.8}& 0.626& 0.730& 0.769& 0.871& 0.842& {\bf 0.906}\\			
			\hline
			{0.8-0.9}& 0.645& 0.778& 0.830& 0.906& 0.881& {\bf 0.926}\\			
			\hline
			{0.9-1}& 0.649& 0.819& 0.884& 0.938& 0.919& {\bf 0.942}\\
			\hline
			\multicolumn{6}{l}{Real Image Noise}\\
			\hline
			{0.5-0.6}& 0.636& 0.770& 0.852& 0.954& 0.915& {\bf 0.958}\\
			\hline
			{0.6-0.7}& 0.635& 0.785& 0.873& 0.963& 0.925& {\bf 0.968}\\
			\hline
			{0.7-0.8}& 0.638& 0.792& 0.881& {\bf 0.971}& 0.932& {\bf 0.971}\\			
			\hline
			{0.8-0.9}& 0.650& 0.807& 0.895& {\bf 0.976}& 0.939& {\bf 0.974}\\			
			\hline
			{0.9-1}& 0.651& 0.811& 0.901& {\bf 0.985}& 0.944& {\bf 0.983}\\	
			\hline
			\multicolumn{6}{l}{Occlusion Noise}\\
			\hline
			{0.5-0.6}& 0.493& 0.408& 0.607& {\bf 0.863}& 0.809& 0.845\\
			\hline
			{0.6-0.7}& 0.540& 0.509& 0.661& {\bf 0.860}& 0.815& 0.852\\
			\hline
			{0.7-0.8}& 0.578& 0.608& 0.717& {\bf 0.872}& 0.827& 0.863\\			
			\hline
			{0.8-0.9}& 0.606& 0.705& 0.781& {\bf 0.886}& 0.844& {\bf 0.885}\\			
			\hline
			{0.9-1}& 0.630& 0.788& 0.867& {\bf 0.951}& 0.908& 0.943\\
			\hline
			\multicolumn{6}{l}{Thresholded Probability Noise}\\
			\hline
			{0.5-0.6}& 0.573& 0.610& 0.731& 0.849& 0.796& {\bf 0.859}\\
			\hline
			{0.6-0.7}& 0.587& 0.673& 0.790& 0.870& 0.833& {\bf 0.878}\\
			\hline
			{0.7-0.8}& 0.604& 0.716& 0.828& {\bf 0.890}& 0.860& 0.887\\			
			\hline
			{0.8-0.9}& 0.627& 0.759& 0.855& {\bf 0.910}& 0.874& 0.897\\	
			\hline
			{0.9-1}& 0.609& 0.806& 0.836& {\bf 0.929}& 0.878& 0.824\\
			\hline
			\multicolumn{6}{l}{Detection Image Noise}\\
			\hline
			{0.5-0.6}& 0.518& 0.528& 0.648& {\bf 0.723}& {\bf 0.703}& {\bf 0.700}\\
			\hline
			{0.6-0.7}& 0.590& 0.631& 0.751& {\bf 0.807}& 0.782& 0.740\\
			\hline
			{0.7-0.8}& 0.590& 0.697& 0.782& {\bf 0.815}& 0.789& 0.772\\			
			\hline
			{0.8-0.9}& 0.653& 0.759& 0.855& {\bf 0.877}& 0.865& 0.852\\
			\hline
			{0.9-1}& 0.671& 0.800& 0.895& {\bf 0.925}& 0.912& {\bf 0.897}\\	
			\hline			
		\end{tabular}
		}
	\end{center}
	\vspace {-10mm}
\end{table}

\vspace{-6mm}
\section{Conclusion}
\label{sec:conclusion}
\vspace{-3mm}

This paper introduced the problem of shape denoising, where the shapes are represented as binary images and the goal is to recover a shape that was deteriorated by a noise process.
The paper introduced six types of noise that could be used to perturb the shapes. Four of the noise types -- real image noise, occlusion noise, thresholded probability noise and detection noise -- are related to real challenges encountered during object segmentation.

The goal of this paper is to provide an objective evaluation of shape modeling methods for object segmentation, independent of the image modality they will be used in applications.

The paper evaluated seven methods from different areas that could be used for shape denoising: 
Active Shape Model (ASM) as a classical segmentation method,
two generative models based on Boltzmann Machines (Deep Boltzmann Machine (DBM) and Convolutional DBM), 
another generative model named Energy Based Model (EBM), 
and three deep-learning based models used for object segmentation: U-Net, DeepLabv3+ and Masked Autoencoder (MAE).

In the future we plan to apply the trained shape models to object segmentation from color images.

\bibliographystyle{IEEEbib}
\bibliography{refs}

\end{document}